\documentclass[10pt,twocolumn,letterpaper]{article}

\usepackage[pagenumbers]{cvpr}     
\usepackage{graphicx}
\usepackage{amsmath}
\usepackage{amssymb}
\usepackage{booktabs}

\usepackage{algorithm}
\usepackage{algorithmic}
\usepackage{multirow}
\usepackage{bbding}

\usepackage{enumitem}

\usepackage{colortbl}
\usepackage{xcolor}

\usepackage[pagebackref,breaklinks,colorlinks]{hyperref}

\usepackage{multirow}

\usepackage[capitalize]{cleveref}
\crefname{section}{Sec.}{Secs.}
\Crefname{section}{Section}{Sections}
\Crefname{table}{Table}{Tables}
\crefname{table}{Tab.}{Tabs.}

\def\eg{\emph{e.g}\onedot}

\newcommand*\samethanks[1][\value{footnote}]{\footnotemark[#1]}

\begin{document}


\title{Beyond Attentive Tokens: Incorporating Token Importance and Diversity for Efficient Vision Transformers}

\author{Sifan Long\textsuperscript{\rm 1,2}\thanks{Equal contribution.}
~~~~Zhen Zhao\textsuperscript{\rm 3,2}\samethanks
~~~~Jimin Pi\textsuperscript{\rm 2}
~~~~Shengsheng Wang\textsuperscript{\rm 1}\thanks{Corresponding authors.}
~~~~Jingdong Wang\textsuperscript{\rm 2}\samethanks \\
\textsuperscript{\rm 1}Jilin University\hspace{16mm}
\textsuperscript{\rm 2}Baidu VIS\hspace{16mm}
\textsuperscript{\rm 3}University of Sydney\\
{\tt\small longsf22@mails.jlu.edu.cn~~zhen.zhao@sydney.edu.au}\\
{\tt\small wss@jlu.edu.cn~~\{pijimin01, wangjingdong\}@baidu.com}
}

\maketitle

\begin{abstract}
Vision transformers have achieved significant improvements on various vision tasks but their quadratic interactions between tokens significantly reduce computational efficiency. 
Many pruning methods have been proposed to remove redundant tokens for efficient vision transformers recently.
However, existing studies mainly focus on the token importance to preserve local attentive tokens but completely ignore the global token diversity. 
In this paper, we emphasize the cruciality of diverse global semantics and propose an efficient token decoupling and merging method that can jointly consider the token importance and diversity for token pruning.
According to the class token attention, we decouple the attentive and inattentive tokens. 
In addition to preserve the most discriminative local tokens, we merge similar inattentive tokens and match homogeneous attentive tokens to maximize the token diversity.
Despite its simplicity, our method obtains a promising trade-off between model complexity and classification accuracy. On DeiT-S, our method reduces the FLOPs by 35\% with only a 0.2\% accuracy drop. Notably, benefiting from maintaining the token diversity, our method can even improve the accuracy of DeiT-T by 0.1\% after reducing its FLOPs by 40\%.

\end{abstract}

\section{Introduction}
\label{sec:intro}
Transformer \cite{vaswani2017attention} has become the most popular architecture in both natural language processing and computer vision communities. Vision transformers (ViTs) \cite{dosovitskiy2020image} have achieved superior performance and outperformed standard CNNs in different vision tasks such as image classification \cite{yuan2021tokens,wu2021cvt,touvron2021training,graham2021levit}, semantic segmentation \cite{xie2021segformer,li2022panoptic,liu2021swin,wang2021pyramid}, and object detection \cite{carion2020end,dai2021up}. The most remarkable advantage of transformer is its ability to effectively capture long-range dependencies between patches in the input image through the self-attention mechanism \cite{rao2021dynamicvit}. However, quadratic interactions between tokens significantly 
degrade the computational efficiency \cite{yang2021nvit}, which motivates many researches on exploring efficient transformers.

\begin{figure}
	\centering
    \includegraphics[width=0.47\textwidth]{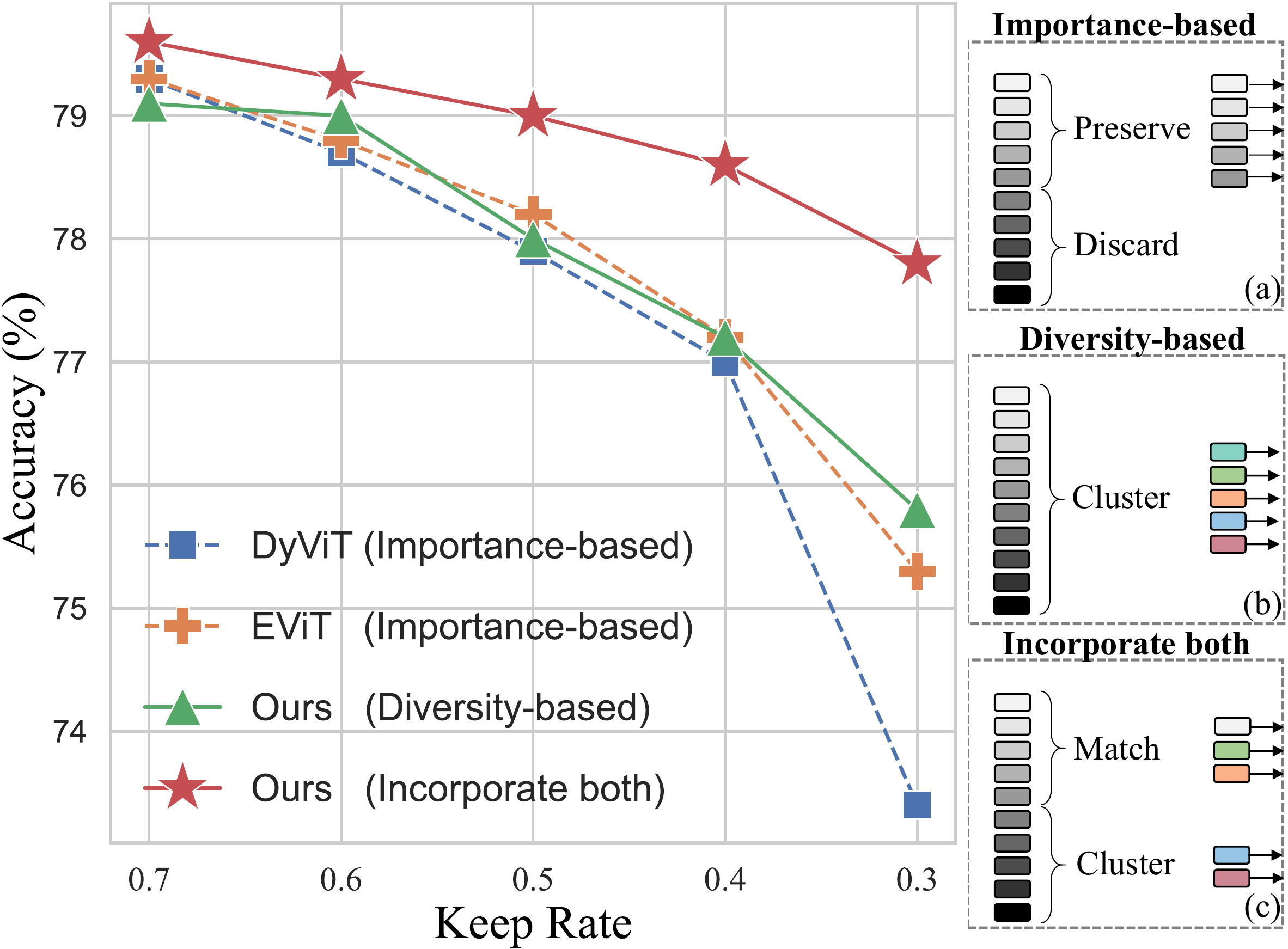}
	\caption{The ImageNet accuracy and keep rate of the pruned DeiT-S. (a) Importance-based method preserves attentive tokens based on the class token attention and masks all inattentive tokens; (b) Diversity-based method clusters similar tokens into a group and then combines tokens from the same group into a new token. (c) Incorporate method decouples and merges tokens to consider token importance and diversity simultaneously.
	}
	\label{FIG:1}
\end{figure}

\begin{figure}
	\centering
	\includegraphics[width=0.47\textwidth]{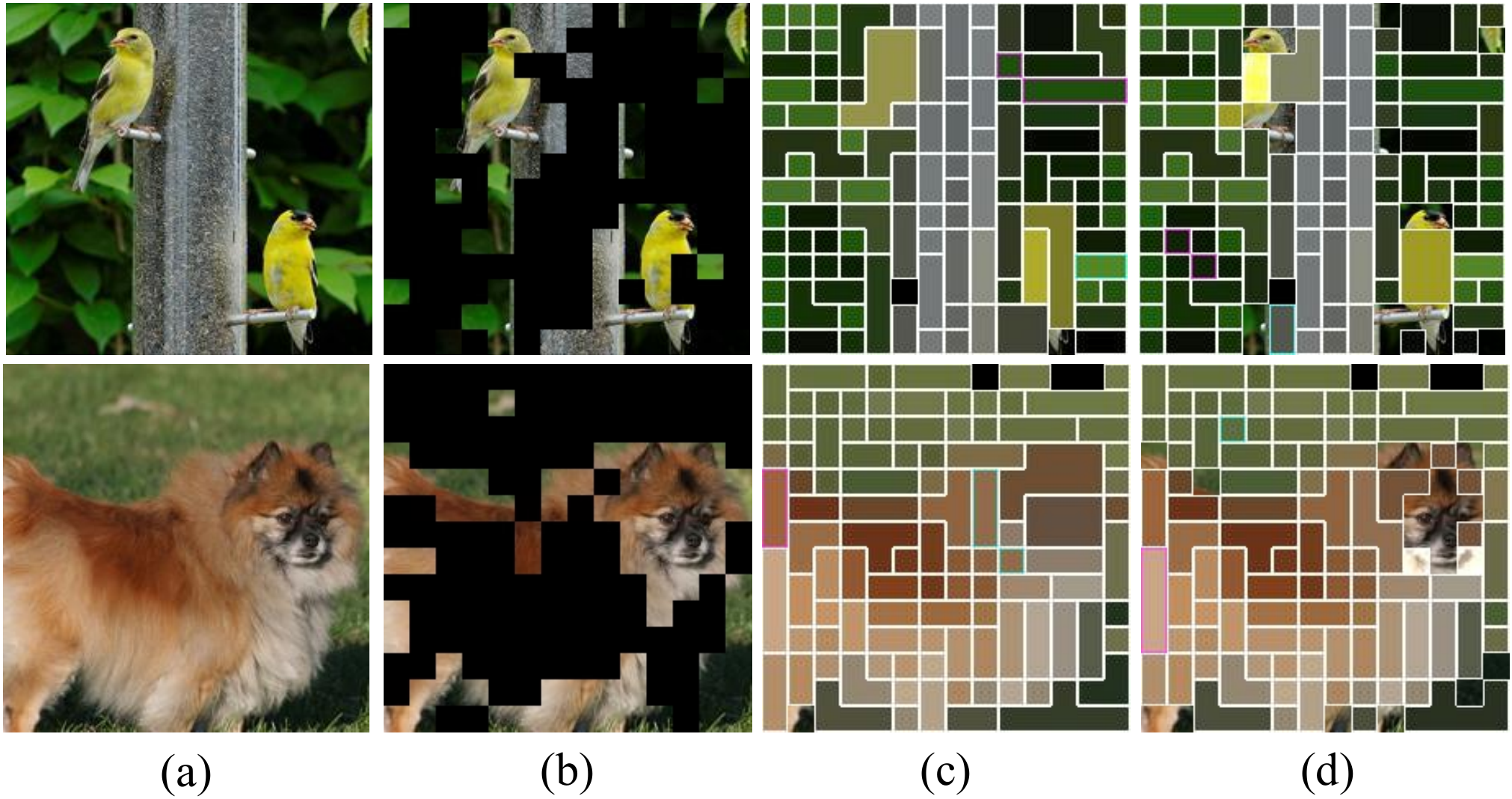}
    \caption{Visualizations of pruning results of different methods on ImageNet with DeiT-S. (a) Original image. (b) Importance-based method masks inattentive tokens. (c) Diversity-based method clusters similar tokens and visualizes the same group of tokens as one colour. 
    (d) Our method preserves the most discriminative tokens, \eg, the heads of birds and dogs. In addition, we merge similar inattentive tokens and match homogeneous attentive tokens, \eg, the grass and leaves.
    }
	\label{FIG:2}
\end{figure}

As one of the most direct and effective ways to reduce computational complexity, token pruning has been widely studied recently. 
Existing studies mainly focus on designing different importance-evaluating strategies to retain attentive tokens and prune inattentive tokens~\cite{rao2021dynamicvit, xu2022evo, pan2021ia, liang2022not, yin2022vit}. In these importance-based works, DyViT \cite{rao2021dynamicvit} introduces an extra module to estimate the importance of each token while EViT \cite{liang2022not} reorganizes image tokens based on the class attention importance score. However, inspired by recent diversity-preserving studies in ViT variants~\cite{han2020ghostnet,han2021transformer,tang2021augmented,ryoo2021tokenlearner,han2022vision,gong2021vision}, we argue that promoting token diversity
is also crucial for token pruning. 
Though inattentive tokens like the image background and low-level textures are not directly related to the classification objects, they can increase the token diversity and improve the expressivity of the model. As discussed in~\cite{xiao2020noise}, image backgrounds (\eg, the grass and leaves in \cref{FIG:2}) can improve the classification accuracy due to their potential relations to foreground objects. To this end, we first investigate a diversity-based pruning strategy on DeiT-S~\cite{touvron2021training} with different keep rates. Specifically, instead of highlighting the token importance, it directly clusters and combines similar tokens into a single one, hereby maximizing the token diversity. Surprisingly, as shown in \cref{FIG:1}, such an intuitive strategy can achieve comparable and even better performance than SOTA importance-based pruning methods, especially at the low keep rate.

Despite its promising performance, the diversity-based strategy cannot retain original attentive tokens and may consequently weaken the discriminative ability of the model. As shown in \cref{FIG:2} (c), the most representative tokens, \eg, eyes and ears of the dog or beaks of two birds, contain critical semantic information for classification tasks but cannot be preserved by the diversity-based strategy. 
To address this issue, we naturally tend to keep all these dominant tokens while maintaining the token diversity, as shown in \cref{FIG:2} (d). In short, a satisfied pruning method should jointly take the token importance and diversity into account, such that the most important local information and the diverse global information can be preserved simultaneously. 

Motivated by these above observations, in this paper, we propose a novel pruning method that incorporates the token importance and diversity  through efficient token decoupling and merging. 
As shown in Figure \ref{FIG:1} (c),
we first decouple the origin token sequence into attentive and inattentive portions based on class token attention.
Instead of discarding inattentive tokens completely, 
we apply a simplified density peak clustering algorithm \cite{rodriguez2014clustering} to efficiently cluster similar inattentive tokens and combine these tokens from the same group into a new one. In addition, unlike existing methods that preserve all attentive tokens, we design a straightforward matching algorithm to fuse homogeneous attentive tokens and improve the calculation efficiency further.
In this way, we can effectively prune tokens while maximizing the preservation of token diversity.
We conduct extensive token pruning experiments to validate the effectiveness of our method. Despite its simplicity, our method achieves superior pruning performance on ImageNet \cite{deng2009imagenet} for two different vision transformers, DeiT \cite{touvron2021training} and LV-ViT \cite{jiang2021all}.
Our main contributions are summarized as follows:
\begin{itemize}
    \item To the best of our knowledge, we are the first to emphasize the token diversity for pruning ViT. We also demonstrate its cruciality through numerical and empirical analysis.
    \item We propose a simple yet effective decoupling and merging method that can simultaneously preserve the most attentive local tokens and diverse global semantics without imposing extra parameters.
    \item Benefiting from incorporating token importance and diversity, our method achieves new SOTA performance on the trade-off between accuracy and FLOPs. It can also be deployed to other token pruning methods, achieving excellent performance improvement.
\end{itemize}

\begin{figure*}
	\centering
	\includegraphics[width=0.9\textwidth]{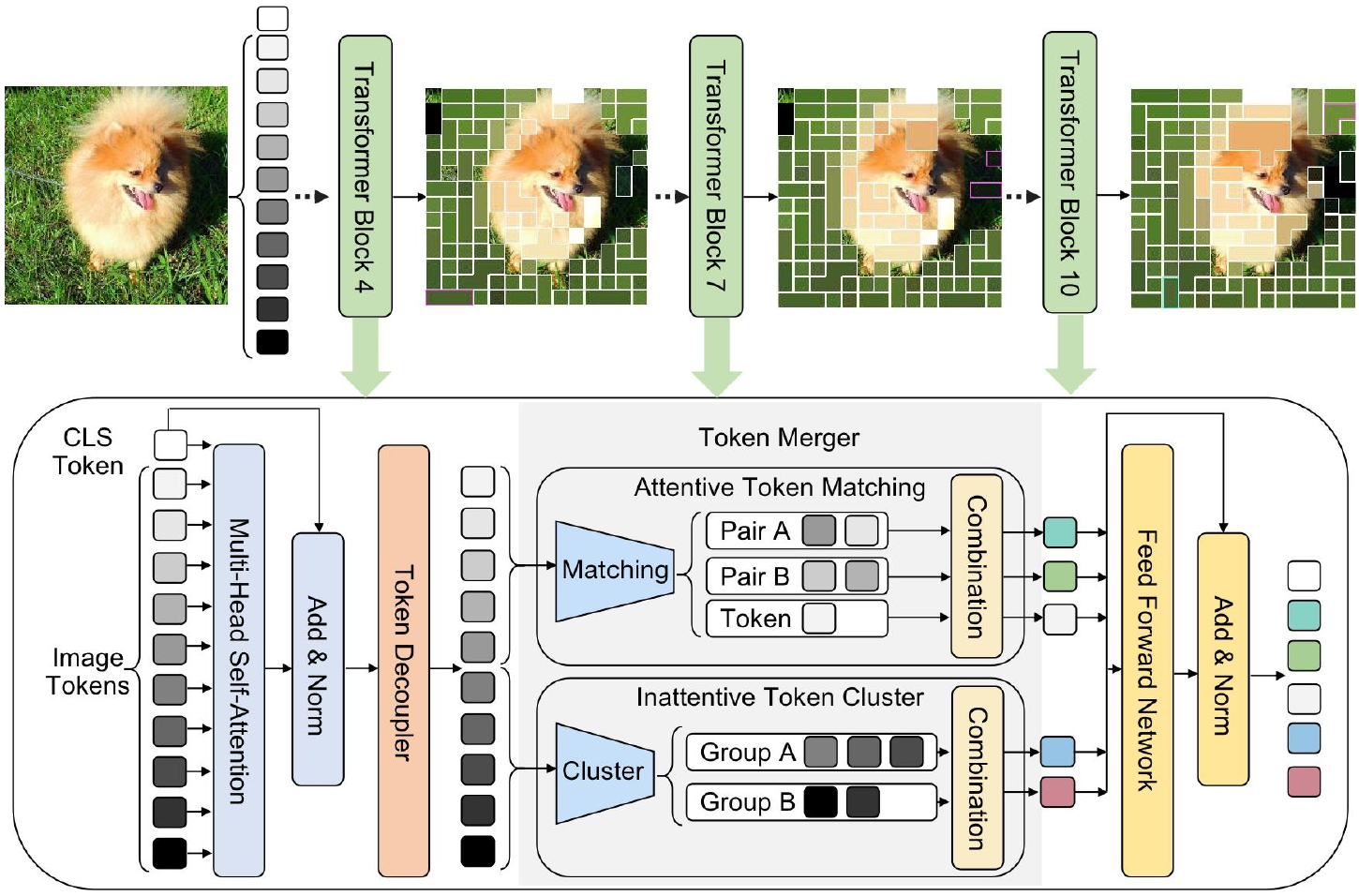}
	\caption{Illustration of our approach. (top) Employ our method at the 4th, 7th, and 10th layers of the DeiT-S model.
	(bottom) Model structure within a single transformer block. We decouple the attentive and inattentive tokens according to class token attention. Then, we cluster inattentive tokens and combine the tokens from the same group into a new token. Meanwhile, we match homogeneous attentive tokens and combine the same pair of tokens.
	}
	\label{FIG:3}
\end{figure*}

\section{Related work}
\label{sec:rwork}
\paragraph{Vision Transformers.} 
Different from convolution networks, the transformer has a significant ability to model long-range dependencies and minimal inductive bias \cite{xu2022evo}. Recent advances suggest that the variants of transformers could be a competitive alternative to CNNs. Visual transformer (ViT) \cite{dosovitskiy2020image} is the first work to apply transformer architecture to achieve STOA performance, but it only replaces the standard convolution in the deep neural network on large-scale image datasets. To free ViT from dependence on large datasets, DeiT \cite{touvron2021training} incorporates an additional token for knowledge distillation to improve the training efficiency of vision transformers. LV-ViT \cite{jiang2021all} further improves the performance by utilizing all the image patch tokens to calculate the training loss intensively. It is equivalent to converting the image classification problem into each token recognition problem. 
While ViT and its follow-ups achieve excellent performance, the complexity quadratic with the number of tokens incurs high computational costs.
Token pruning aims to reduce redundant tokens and improve the inference efficiency of various ViT backbones.

\paragraph{ViT Token pruning.} 
Though ViT has achieved competitive accuracy in vision tasks \cite{yuan2021tokens,wu2021cvt,touvron2021training,graham2021levit,carion2020end,dai2021up}, it needs huge memory and computational resources. Therefore, how to build a more efficient transformer draws researchers’ interest. Compared with CNN, the higher computing cost of the transformers is mainly due to the quadratic time complexity of multi-head self-attention (MHSA). Accordingly, some work \cite{rao2021dynamicvit,pan2021ia,liang2022not,xu2022evo} attempts to prune tokens based on importance score in transformer. Based on whether extra parameters need to be introduced to the model, we divide the existing token pruning methods into the following two groups. One group performs token pruning by inserting prediction modules. DyViT \cite{rao2021dynamicvit} designs a lightweight prediction module inserted into different layers to estimate the importance score of each token to prune redundant tokens given the current features. IA-RED2 \cite{pan2021ia} introduces interpretable modules to dynamically delete redundant patches that are not related to the input. AdaViT \cite{meng2022adavit} connects a lightweight decision network to the backbone to dynamically generate decisions. The other group leverages the class token attention to keep attentive tokens. EViT \cite{liang2022not} divides image tokens into attentive and inattentive tokens according to class token attention, retains attentive tokens and discards inattentive image tokens to reorganize image tokens. Evo-ViT \cite{xu2022evo} distinguishes informative and uninformative tokens through global class attention for slow and fast updates, respectively. A-ViT \cite{yin2022vit} designs an adaptive token pruning mechanism based on class token attention, which dynamically adjusts the calculation cost of images with different complexity. Unlike these token pruning methods, which only focus on the importance of tokens, our method also considers the diversity of token semantic information. Therefore our method achieve incredible performance.

\section{Preliminaries}
In standard vision transformers \cite{dosovitskiy2020image}, each input image $\mathrm{I} \in \mathbb R^{H \times W \times C}$ is first converted into a single-dimensional patch sequence $\mathrm{X} \in \mathbb R^{N \times P^{2} \times C}$. Then all patches are mapped into $D$-dimensional token embeddings via a trainable linear layer. Additionally, a learnable position embedding $\mathrm{E}_{\text {pos}} \in \mathbb{R}^{(N+1)\times D}$ is added to token embedding to retain position information. Formally, the input patch sequence can be represented as:

\begin{equation}
\mathrm{X}=\left[\mathrm{x}_\mathrm{cls} ; \mathrm{x}_{1} ; \ldots ; \mathrm{x}_\mathrm{N}\right]+\mathrm{E}_{\text {pos}},
\end{equation}
where $\mathrm{x}_{\text {cls}}$ denotes the learnable class token that serves as the image representation, and $\mathrm{x}_{i}$ denotes the token of the $i$-th patch with $i \geq 0$. Afterwards, such token sequence is fed into a ViT model with $L$ stacked transformer blocks, each of which consists of a multi-head self-attention (MHSA) module and a feed forward network (FFN).

\subsection{MHSA \& FFN}In MHSA, the input token sequence is linearly mapped into three different matrices of query $\mathrm{Q}$, key $\mathrm{K}$, and value $\mathrm{V}$, respectively. MHSA can be formulated as:
\begin{align} 
\operatorname{MHSA}\left(\mathrm{Z}\right)=\text{Concat}\left[\operatorname{softmax}\left(\frac{\mathrm{Q}^{\mathrm{h}}\left(\mathrm{K}^{\mathrm{h}}\right)^{\top}}{\sqrt{\mathrm{d}}}\right)\mathrm{V}^{\mathrm{h}}\right]_{\mathrm{h}=1}^{\mathrm{H}},
\end{align}
where $\mathrm{Z}$ is the token sequence of $\mathrm{N+1}$ tokens. Concat$[\cdot]$ outputs the feature concatenation of $\mathrm{H}$ heads. $\mathrm{Q}^\mathrm{h}$, $\mathrm{K}^\mathrm{h}$ and $\mathrm{V}^\mathrm{h}$ are projection matrices of $\mathrm{Q}$, $\mathrm{K}$, and $\mathrm{V}$ in the $\mathrm{h}$-th head, respectively. $\mathrm{d}$ is the feature dimension of the single head.

FFN typically consists of two fully-connected layers and a nonlinear mapping layer, which can expressed as:
\begin{equation}
\operatorname{FFN}\left(\mathrm{Z}\right)=\operatorname{Sigmoid}(\operatorname{Linear}(\operatorname{GeLU}(\operatorname{Linear}(\mathrm{Z})))),
\end{equation}
where $\operatorname{Linear}$ denotes the fully-connected layers and $\operatorname{GeLU}$ is an non-linear activation function.

\subsection{Computation Complexity}The dimension of the input token sequence is ${N \times D}$, where $N$ is the number of tokens and $D$ is the embedding dimension of each token. Thus the calculational costs of MSHA and FFN modules are $\mathcal{O}\left(4 N D^{2}+2 N^{2} D\right)$ and $\mathcal{O}\left(8ND^{2}\right)$, respectively. Obviously, vision transformers require very intensive computational costs, with the total computational complexity of $\mathcal{O}\left(12 N D^{2}+2 N^{2} D\right)$. Since reducing the channel dimension $D$ only affects the calculation of current matrix multiplication, most related works tend to prune tokens, \eg reducing the number of $N$, to reduce all operations linearly or even quadratically. 


\section{Methodology}
\subsection{Overview}
Different from existing works only focus on attentive tokens, our method incorporating token importance and diversity to obtain efficient and accurate vision transformers. To this end, we propose the token decoupling and merging method, achieving promising trade-offs between the FLOPs and accuracy. As shown in Figure \ref{FIG:3}, we insert our approach at the 4th, 7th, and 10th layers of the DeiT-S model. The approach has two main components: \textbf{the token decoupler} and \textbf{the token merger}. The decoupler divides the origin token sequence into attentive and inattentive sections based on class token attention. Then the merger clusters similar inattentive tokens and matches homogeneous attentive tokens. In this section, we first demonstrate how preserving token diversity benefits token pruning and then present the two main components in detail.

\begin{figure}
	\centering
	\includegraphics[width=0.5\textwidth]{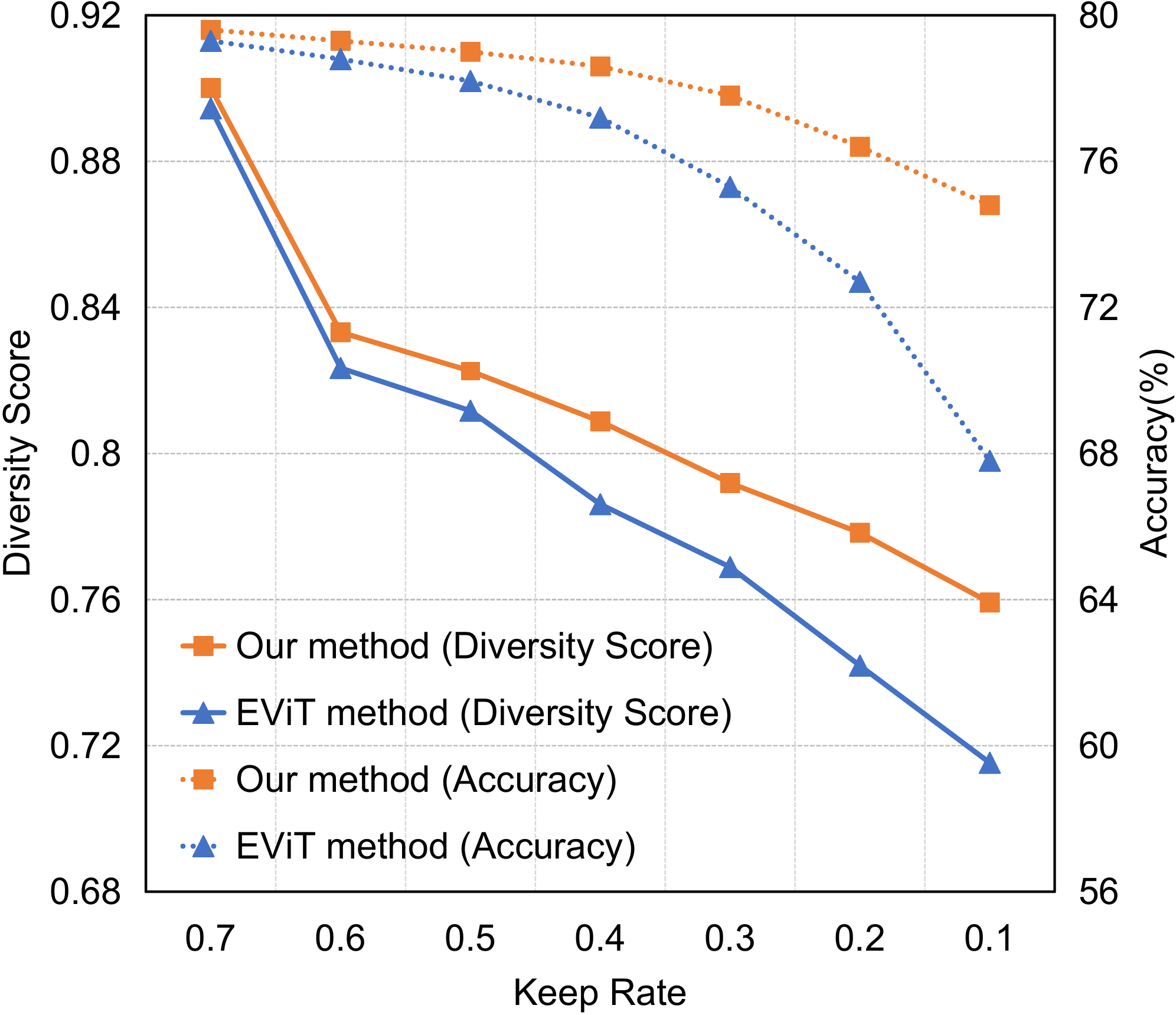}
    \caption{Comparing the pruning results of our method and the EViT method with different keep rates on DEiT-S in terms of the diversity score and classification accuracy on ImageNet. The diversity score is obtained at the final pruning layer.}
	\label{FIG:4}
\end{figure}

\subsection{Token diversity matters}
In the literature, most work only emphasize retaining important tokens but directly discard all the remaining ones to achieve satisfactory token keep rates. However, inspired by the observations in \cite{xiao2020noise}, that even the image background can help improve foreground-instance classification, we argue that the less important tokens could also contain useful semantic information and be an effective complementary to the information diversity. Also, as discussed in \cite{han2020ghostnet,han2021transformer,tang2021augmented,ryoo2021tokenlearner,han2022vision,gong2021vision}, 
the token diversity is very critical to optimize transformer structures. Therefore, appropriately preserving these inattentive tokens augments the diversity of semantic information, which can be beneficial to token pruning. On the contrary, blindly discarding tokens will cause irreversible loss of semantic information, especially at the low keep rates. Referring to \cite{dong2021attention,han2022vision,tang2021augmented,tang2022patch}, we leverage the difference between the token and a rank-1 matrix to measure the diversity of token sequence $\mathrm{Z}$. The diversity scores $r\left(\mathrm{Z}\right)$ can be calculated as:
\begin{equation}
r\left(\mathrm{Z}\right)=\left\|\mathrm{Z}-1 z^{\top}\right\|,\text { where } \boldsymbol{z}=\operatorname{argmin}_{\boldsymbol{z}^{\prime}}\left\|\mathrm{Z}-1 z^{\prime \top}\right\|,
\end{equation}
where $\|\cdot\|$ represents $l_1$ norm. $\mathrm{Z} \in \mathbb{R}^{N \times C}$ is the token sequence of $N$ tokens and $\boldsymbol{z},\boldsymbol{z}^{\prime} \in \mathbb{R}^{C}$ is one of the tokens. $z^{\top}$ is the matrix transpose of $\boldsymbol{z}$ and 1 is an all-ones vector. The rank of matrix $1 z^{\top}$ is 1. A larger diversity score indicates a more diverse token sequence.

We investigate how the diversity scores affect the token pruning performance. In Figure \ref{FIG:4}, we examine the classification accuracy and the diversity score at different keep rates. Obviously, we can see that the token sequence diversity score is positively correlated with classification accuracy. In either the EViT method or our proposed method, higher diversity score consistently result in higher accuracy. In addition, it can also be observed that, as for the EViT method, the diversity score and classification accuracy drop rapidly as the keep rate decreases. Differently, benefiting from our diversity-preserving token merging strategy, our method can maintain relatively higher diversity scores at different keep-rates and thus consistently outperform the EViT method, especially at the low keep-rate. Therefore, maintaining higher token diversity is crucial to improve classification accuracy.

\subsection{Token Decoupler}
In order to fully consider token importance while maintaining diversity, we prioritize the attentive tokens to preserve the most important semantic information. 
Therefore, we decouple original token sequence into attentive and inattentive tokens so that we maintain token diversity and importance simultaneously. 
Same as \cite{vaswani2017attention}, we split the tokens into two groups by comparing their similarities with the class tokens. 
Mathematically, the similarity scores $\mathrm{Attn}_{\mathrm{cls}}$ between the class token and other tokens as class token attention can be calculated by

\begin{equation}
\mathrm{Attn}_{\mathrm{cls}}=\operatorname{Softmax}\left(\frac{\mathrm{q}_{\text {cls }} \cdot \mathrm{K}^{\top}}{\sqrt{\mathrm{d}}}\right),
\end{equation}
where ${\mathrm{q}_\mathrm{cls}}$ denotes the class token of query vector.
With $N$ tokens in total and the keep rate of $\eta$, we choose the top-K tokens as attentive tokens according to attention scores. The remaining $N$-$K$ tokens are identified as inattentive tokens that contain less information. Moreover, in the multi-head self-attention layer, we calculate the average of the attention scores of all heads.


\begin{table*}[ht]
    \centering
	\begin{tabular}{l|c|ccccc}
		\toprule
		Model& Method& Top-1 Acc. (\%) & Params (M) & FLOPs (G) & FLOPs ↓(\%) & Throughput (img/s)\\
		\midrule 
        \multirow{7}{*}{DeiT-T} & DeiT-T \cite{touvron2021training}&	72.2 &5.6 &1.3  & 0.0 & 2536\\
        &DyViT \cite{rao2021dynamicvit}&71.2 &5.9 &0.9 & 30.8 & 3542\\
        &PS-ViT \cite{tang2022patch}&72.0 &5.6 &0.9 & 30.8 & 3563\\
        &SViTE \cite{chen2021chasing}&70.1 &4.2 &0.9 & 30.8 & 2836\\
        &SPViT \cite{kong2021spvit}&72.2 &5.6 &	1.0  & 23.1& -\\
        &Evo-ViT \cite{xu2022evo}& 72.0 &5.6 &0.8  & 38.5 & 3627\\
        \rowcolor{green!20}
        &\textbf{Ours-DeiT-T}& 72.3 &5.6 & 0.8 & 38.5 &3641\\
        \hline
        \multirow{12}{*}{DeiT-S} & DeiT-S\cite{touvron2021training}&   79.8 &22.1  &4.6 & 0.0 	&943\\
        &DyViT\cite{rao2021dynamicvit}&79.3 &22.8 &  2.9 & 37.0 &1420\\
        &PS-ViT\cite{tang2022patch}& 79.4 &22.1 &2.6 & 43.5	&1392\\
        &IA-RED2\cite{pan2021ia}& 79.1 &22.1 &3.2 & 30.4	&1362\\
        &Evo-ViT\cite{xu2022evo}& 79.4 &22.1 &3.0 & 34.8	&1449\\
        &EViT\cite{liang2022not}& 79.5  &22.1 &3.0 & 34.8	&1455\\
        &A-ViT\cite{yin2022vit}& 78.6  &22.1 &3.6	 & 21.7 &-\\
        \rowcolor{green!20}
        &\textbf{Ours-DeiT-S}& 79.6 &22.1 &3.0 & 34.8 &1468\\
        \rowcolor{green!20}
        &\textbf{EViT+Ours}& 79.6 &22.1 & 3.0 & 34.8 &1459\\
        \hline
        \multirow{7}{*}{DeiT-B} & DeiT-B \cite{touvron2021training}&	81.8 & 86.6 &	17.6 & 0.0 &302\\
        &DyViT\cite{rao2021dynamicvit}&	 81.3 &- &	11.6 & 34.1 &454\\
        &PS-ViT\cite{tang2022patch}&81.5 &86.6 &	11.6 & 34.1&445\\
        &IA-RED2\cite{pan2021ia}&80.9 &86.6 &	11.6 & 34.1&453\\
        &Evo-ViT\cite{xu2022evo}&81.3 &86.6 &	11.6 & 34.1&448\\
        &EViT\cite{liang2022not}&81.3 &86.6 & 	11.6 & 34.1&450\\
        \rowcolor{green!20}
        &\textbf{Ours-DeiT-B}&	82.0 &86.6 &	11.6 & 34.1 &462\\
		\bottomrule
	\end{tabular}
	\caption{Comparisons with existing token pruning methods on DeiT. We report the top-1 classification accuracy, FLOPs, and throughput on ImageNet. ‘FLOPs ↓’ denotes the reduction ratio of FLOPs. 
	}
	\label{tab1}
\end{table*}

\begin{table}[t]
    \centering
    \begin{tabular}{lccc}
		\toprule
		Method&  Top-1 Acc    & FLOPs & Params\\
		&(\%)&(G)&(M)\\
		\midrule 
        ViT-Base/16 \cite{dosovitskiy2020image}	 &77.9	 &17.6	 &86.6\\
        
        DeiT-S \cite{touvron2021training}	&79.8	 &4.6	&22.1	 \\
        DeiT-Base/16 \cite{touvron2021training}	 &81.8	 &17.6	 &86.6\\
        
        PVT-Small \cite{wang2021pyramid}	&79.8	 &3.8	&24.5	 \\
        PVT-Medium \cite{wang2021pyramid}	 &81.2	 &6.7	 &44.2	 \\
        CPVT-Small-GAP \cite{chu2021conditional}	 &81.5	 &4.6	 &23.0\\
        
        CoaT Mini \cite{xu2021co}	&80.8	&6.8	 &10.0	 \\
        CoaT-Lite Small \cite{xu2021co}	 &81.9	 & 4.0	 &20.0\\
        
        CrossViT-S \cite{chen2021crossvit}	&81.0	&5.6	&26.7	 \\
        CrossViT-B \cite{chen2021crossvit}	 &82.3	 &  14.1	 & 64.1\\
        
        Swin-T \cite{liu2021swin}	&81.3	 &4.5	 &29.0	 \\
        Swin-S \cite{liu2021swin}	 &83.0	 & 8.7	 &50.0\\
        Swin-B \cite{liu2021swin}	 &83.3	 &  15.4	 &  88.0\\
        
        T2T-ViT-14 \cite{yuan2021tokens}	 &81.5	 &5.2	 &22.0	\\
        T2T-ViT-19 \cite{yuan2021tokens}	 &81.9	 &8.9	 &39.2\\
        T2T-ViT-24 \cite{chu2021conditional}	 &82.2	 & 21.2	 &104.7\\
        
        RegNetY-8G \cite{radosavovic2020designing}	 &81.7	 &  8.0	 &  39.0\\
        RegNetY-16G \cite{radosavovic2020designing}	 &82.9	 &  16.0	 & 84.0\\
        
        LV-ViT-S \cite{jiang2021token} & 83.3&	6.6 &26.2\\
        
        \midrule
        DyViT-LV-S&	83.0&	4.6&26.2\\
        EViT-LV-S&	83.0&	4.7&26.2\\
        \rowcolor{green!20}
        \textbf{Ours-LV-S}&\textbf{83.1}&\textbf{4.7}&26.2\\
		\bottomrule
	\end{tabular}
	\caption{ 
    Comparisons with state-of-the-art vision transformers on ImageNet. We prune the LV-ViT-S as the base model.
    }
	\label{tab2}
\end{table}

\subsection{Token Merger}
We apply different strategies for attentive and inattentive tokens when merging tokens. For inattentive tokens, we first apply density peak clustering algorithm to cluster inattentive tokens and then combine the tokens from the same group into new token by weighted sum. In this way, we can integrate a new inattentive token sequence $\mathrm{T}=\left[\mathrm{t}_{1} ; \ldots ; \mathrm{t}_\mathrm{n}\right]$. For attentive tokens, we adapt a straightforward matching algorithm to fuse homogeneous attentive tokens. The fused token sequence is $\mathrm{P}=\left[\mathrm{p}_{1} ; \ldots ; \mathrm{p}_\mathrm{m}\right]$. We concat $\mathrm{T}$ and $\mathrm{P}$ to obtain the pruned token sequence ${\mathrm{Z}}=\left[\mathrm{z}_\mathrm{cls} ; \mathrm{p}_{1} ; \ldots ; \mathrm{p}_\mathrm{m}; \mathrm{t}_{1} ; \ldots ; \mathrm{t}_\mathrm{n}\right]$.

\paragraph{Inattentive Token Clustering.}
Common K-means clustering algorithm requires multiple iterations to obtain satisfactory results, reducing throughput in practice and defeating the intent of speeding up the model.
After research, we found that the density clustering algorithm can quickly find classes of arbitrary shape by exploiting the density connectivity of classes. Therefore, we simplify an efficient density peak clustering algorithm (DPC) with neither an iterative process nor more parameters.
We follow the DPC algorithm proposed in \cite{rodriguez2014clustering}. It assumes that the cluster center is surrounded by low-density neighbours, and that the distance between the cluster center and any high-density points is relatively large. We calculate two variables for each token $i$: the density $\rho$ and the minimum distance from the higher density token $\delta$. Given a set of tokens x, we calculate the density of each token $\rho$ by
\begin{equation}
\rho_{i}=\exp \left( \sum_{z_{j} \in\mathrm{Z}}\left\|z_{i}-z_{j}\right\|_{2}^{2}\right).
\end{equation}
where $\mathrm{Z}$ denotes the set of tokens. $z_{i}$ and $z_{j}$ are corresponding token features.

For the token with the highest density, its minimum distance is set to the maximum distance between it and any other tokens. We define $\delta_{i}$ as the minimum distance between the token $i$ and any other token with higher density. The minimum distance of each token is:
\begin{equation}
\delta_{i}=\left\{\begin{array}{l}
\min _{j: \rho_{j}>\rho_{i}}\left\|z_{i}-z_{j}\right\|_{2}, \text { if } \exists j \text { s.t. } \rho_{j}>\rho_{i} \\
\max _{j}\left\|z_{i}-z_{j}\right\|_{2}, \text { otherwise }
\end{array}\right..
\end{equation}

We denote the clustering center score of the $i$-th token as $\rho_{i} \times \delta_{i}$.  Higher scores mean higher potential to be cluster centers. We select top-K-scored tokens as cluster centers. The DPC algorithm assigns each remaining token to the cluster whose cluster center is closest to the token and has a higher density. 

\begin{figure*}
	\centering
	\includegraphics[width=1\textwidth]{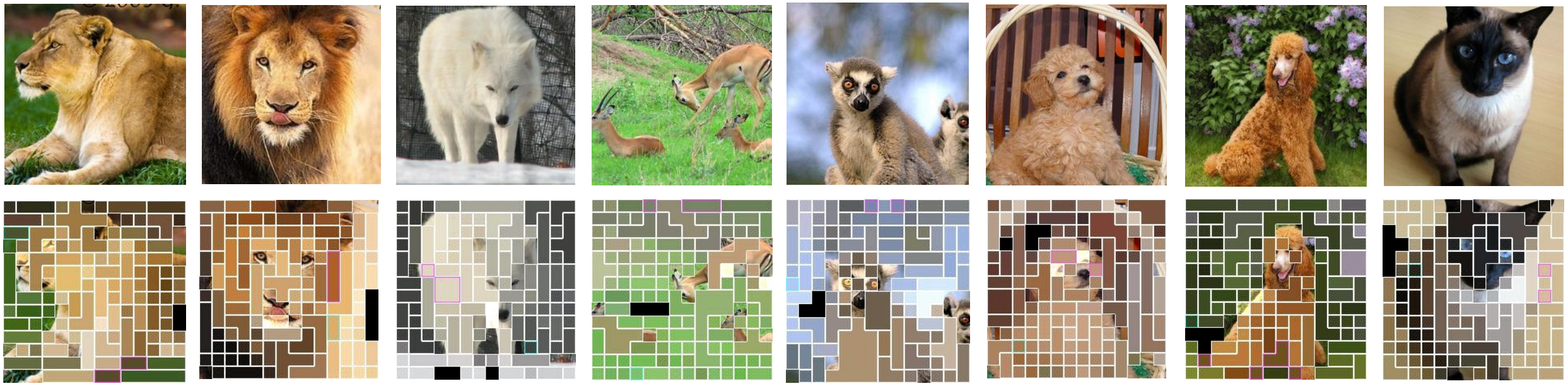}
	\caption{Visualization of token merger results on DeiT-S. The masked areas of different colours represent the inattentive tokens are divided into dissimilar token groups. Our method clusters similar inattentive tokens into a group and matches homogeneous attentive tokens. We visualize the same groups/pairs of tokens as the same colour.
    }
	\label{FIG:5}
\end{figure*}

\paragraph{Attentive Token Matching.}
See example images in Figure \ref{FIG:5}. Homogeneous tokens are also present in foreground objects (attentive tokens), such as the cheeks of animals. 
This redundancy makes it possible to fuse homogeneous attentive tokens to reduce the number of tokens while maintaining accuracy. We could apply the same token clustering strategy as did for inattentive tokens. However, since the attentive tokens contain critical semantic information for the final classification task, it would be best if we can preserve the original tokens.
To address this problem, we customize a straightforward matching algorithm that keeps the most important tokens while fusing homogeneous tokens.
Specifically, we define the cosine similarity metric to determine the similarity between different tokens and calculate cosine similarity scores between attentive tokens. Then we select top-K most similar token pairs as homogeneous tokens. Finally, we combine each pair of tokens into a new token and contact the remaining attentive tokens.

Although tokens in the same set have similar semantic information, the semantic importance of each token is not necessarily the same. Instead of blindly averaging the tokens in the same set, we combine these tokens by a weighted sum. By introducing a class token attention to represent the importance, we combine the same set of tokens into a new token by
\begin{equation}
p_i= \sum_{j \in C_{i}} {s_{j}} z_{j},
\end{equation}
where $s_{j}$  denotes the importance score of token $z_j$, and $C_i$ denotes the $i$-th set. 

\section{Experiments}
\subsection{Setup}
\paragraph{Dataset and evaluation metrics.} Our experiments are conducted on ImageNet-1K \cite{deng2009imagenet} with 1.28 million training images and 50000 validation images. We report the top-1 classification accuracy and the floating-point operations (FLOPs) to evaluate model efficiency.  In addition, we measure the throughput of the models on a single NVIDIA V100 GPU with batch size fixed to 256.

\paragraph{Implementation details.}
To demonstrate the generalization of our method, we conduct token pruning on different ViT models including DeiT-T, DeiT-S, DeiT-B~\cite{touvron2021training}, and LV-ViT-S \cite{jiang2021all}. Following \cite{liang2022not},  we employ our method at the 
$4^{th}$, $7^{th}$, and $10^{th}$ layers of the DeiT-T/S/B model and at the $4^{th}$, $8^{th}$, and $12^{th}$ layers for LV-ViT-S \cite{jiang2021all}.
We utilize the same training settings as the original papers of DeiT \cite{touvron2021training} and LV-ViT \cite{jiang2021all}. Following~\cite{yuan2021tokens}, we incorporate a cosine scheduler into our learning strategy where the keep-rate gradually decreases from 1 to the target value for 100 epochs. For fair comparisons, all of our models are trained from scratch for 300 epochs on 8 NVIDIA V100.

\subsection{Main Results}
\paragraph{Comparisons with the-state-of-the-arts.}
We compare our method with SOTA token pruning methods, results are shown in Table \ref{tab1}.
We leveraged the $\eta$ indicates the keep rate. We report the top-1 accuracy, FLOPs, and throughput for each model. Compared to previous work, our method achieves new SOTA performance with similar computation costs. Specifically, on the classic model DeiT-S \cite{touvron2021training}, the top-1 accuracy degradation of our pruned models is controlled within 0.2\% when the computation costs decreases by 35\%. In addition, the superiority of our method is more obvious at lower keep-rates. When the compression ratio of DeiT-S is increased to 50\%, we improve 0.5\% compared to the best counterpart. In particular, the compression ratio of our method is close to 40\% on the DeiT-T \cite{touvron2021training} model, and the accuracy is even better than the original model. Owing to the token diversity and importance are orthogonal for token pruning, our method can be plugged into EViT and achieve a performance improvement of 0.1\%.
As shown in Table \ref{tab2}, we further conduct experiments on the  deep-narrow transformer LV-ViT \cite{jiang2021all}. We observe that our method achieves better accuracy-complexity trade-offs on different keep rates compared to the current foremost CNN and ViT architectures. 
\begin{figure}
	\centering
	\includegraphics[width=0.47\textwidth]{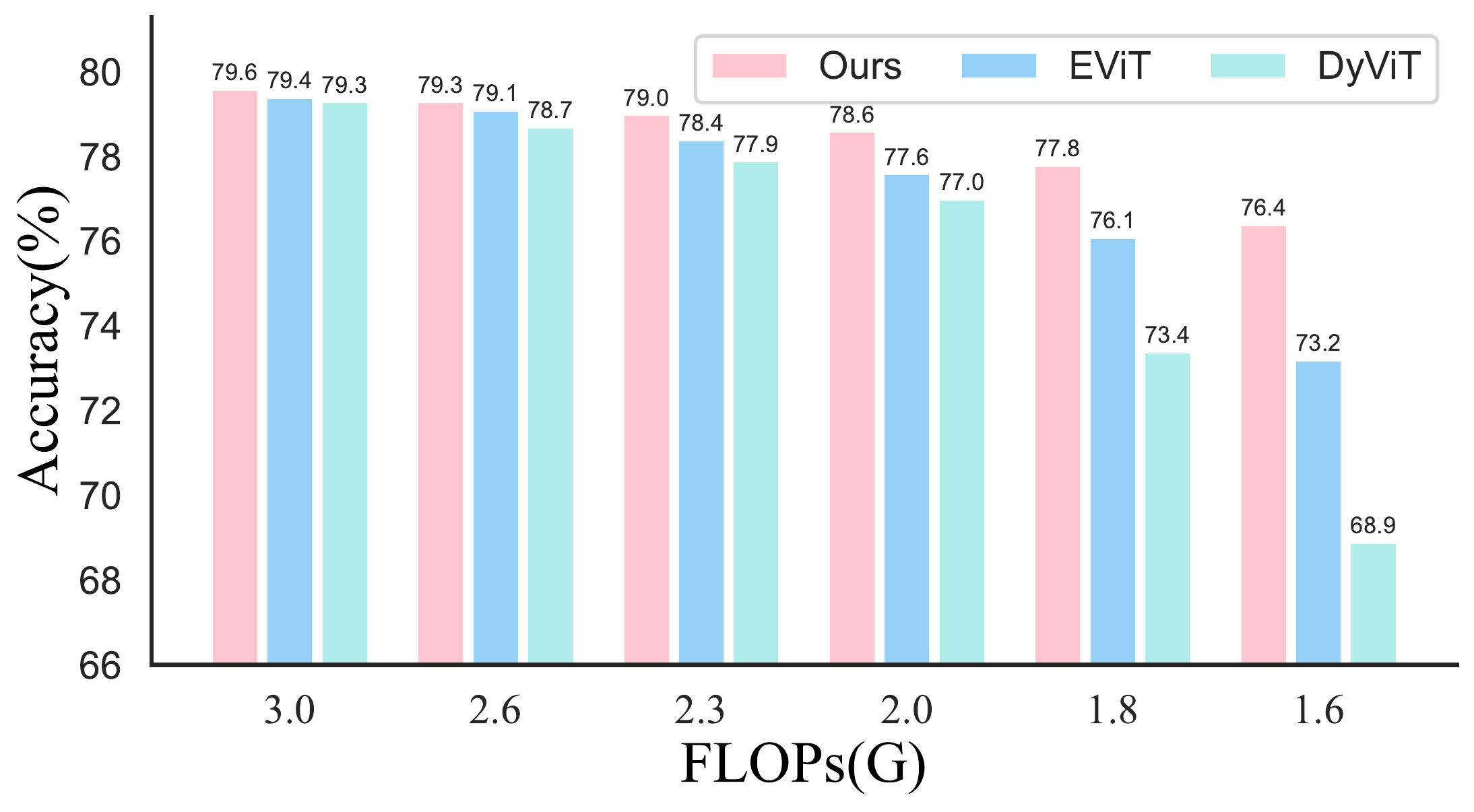}
	\caption{Performance comparisons of DyViT, EViT, and our method with different FLOPs.}
	\label{FIG:7}
\end{figure}

\paragraph{Performance of existing methods on each keep rate.}
As shown in Figure \ref{FIG:7}, our method consistently achieves the best performance while the other two methods obtain close
performance. 
In addition, with the decrease of keep rate, the classification accuracy of existing token pruning methods drops sharply.  
Fortunately, our method alleviates the phenomenon by preserving the diversity of token semantic information.
Especially when the FLOPs of DyViT is reduced to 1.6G, the classification accuracy drops by more than 10\%. 
This is because completely discarding inattentive tokens greatly decreases token diversity, resulting in the reduction of the semantic information of the original token sequence. 
We apply the token decoupling and merging to simultaneously consider the token importance and diversity, achieving incredible accuracy at low keep rates. 
Intuitively, when we only keep a few tokens, merging tokens obviously makes more sense than keeping only top-K attentive tokens.


\paragraph{Visualization of the token merger results.}
To further show the interpretability of our method, we visualize the final token merger results back to its original input patches in Figure \ref{FIG:5}.
We notice that our method pays attention to the regions that contribute more to image prediction instead of uninformative backgrounds. \eg, the animal’s five sense organs are preserved. It demonstrates that our method effectively decouple the attentive and inattentive tokens. Unlike other methods that mask all inattentive tokens, our method combines background patches with similar semantics. \eg, the animal’s fur is merged into a single token. It implies that our method not only focuses on attentive tokens but also maintains the diversity of token semantics.  It is also worth pointing out that the paired eyes are matched and combined into a token in the fifth and sixth image. It indicates that our method effectively fuse homogeneous attentive tokens and reduces the potential redundancy.

\begin{table}[t]
    \centering
		\begin{tabular*}{\hsize}{@{}@{\extracolsep{\fill}}lccc@{}}
			\toprule
			Strategy &	Acc (\%)&	FLOPs (G)\\
			\midrule 
            Deit-S&	79.8&	4.6\\
            \hline
            \multicolumn{3}{c}{DeiT-S/$\eta$=0.7}\\
            \hline
            + attentive token preservation&	79.3&	3.0\\
            + inattentive token pack&	79.3&	3.0\\
            + inattentive token clustering&	79.5&	3.0\\
            + attentive token matching&	79.6&	3.0\\
            \hline
            \multicolumn{3}{c}{DeiT-S/$\eta$=0.5}\\
            \hline
            + attentive token preservation&	78.2&	2.3\\
            + inattentive token pack&	78.4&	2.3\\
            + inattentive token clustering&	78.8&	2.3\\
            + attentive token matching&	79.0&	2.3\\
			\bottomrule
	\end{tabular*}
	\caption{Ablation study on our method with different keep-rate $\eta$.} 
	\label{tab3}
\end{table}

\subsection{Ablation Analysis}
\paragraph{Effectiveness of each module.}
As shown in Table \ref{tab3}, we add the sub-modules one by one to evaluate the effectiveness of each module.
i) Attentive token preservation. Discard inattentive tokens based on the class token attention in pruning layers; ii) Inattentive token pack. Pack all inattentive tokens into one token; iii) Inattentive token clustering. Cluster inattentive tokens and combine the tokens of the same group into a new token; iv) Attentive token matching. Match attentive tokens and combine the tokens of the same pair into a new token; 
It is evident that since the clustering module preserves token diversity, the accuracy is improved by 0.2\% and 0.8\% at keep rates of 0.7 and 0.5, respectively.
Noteworthy, the lower keep rates, the better our method works. In addition, the attentive token matching module further reduces the FLOPs of the model while maintaining accuracy by fusing homogeneous attentive tokens. 

\begin{table}[t]
    \centering
	\begin{tabular*}{\hsize}{@{}@{\extracolsep{\fill}}lccc@{}}
		\toprule
		Method &	Acc&	FLOPs& Throughput\\
		       &    (\%)&	 (G) &(img/s)\\
		\midrule 
		\multicolumn{4}{c}{Pooling strategy}\\
		\hline
		Average pooling&	78.1&	2.3& 1630\\
        Max pooling&    	78.1&	2.3& 1623\\
        Spatial pooling&	78.2&	2.3& 1605\\
        \hline
        \multicolumn{4}{c}{Sub-sampling strategy}\\
		\hline
        Convolution&	    78.2&	2.3& 1454\\
        MLP&	            78.3&	2.3& 1447\\
        \hline
        \multicolumn{4}{c}{Cluster strategy}\\
		\hline
		K-means(1 iter)&	        78.6&	2.3& 1386\\
        K-means(3 iter)&	        78.8&	2.3& 1231\\
        Ours&	            79.0&	2.3& 1670\\
		\bottomrule
	\end{tabular*}
	\caption{Different token merger strategies on DeiT-S.}
	\label{tab4}
\end{table}
\paragraph{Different Token Merger Strategy.}
As presented in Table \ref{tab4}, we compare several common token merging strategies to assess the effectiveness of our approach. i) Pooling strategy. Utilize the pooling operation to reduce the number of tokens.  ii) Sub-sampling strategy. Adding a series of convolution layers between MHSA and FFN to decrease the token dimension. iii) Cluster strategy. Cluster the tokens and combine the tokens of the same group into a new token. 
We observe that the clustering strategy generally improves the accuracy by 0.4\% compared to other token merging strategies.
A possible reason is that the clustering strategy obtains more inductive bias at the same computational cost. 
However, we find that the throughput of K-means algorithm is lower, indicating that it does not perform well in terms of model acceleration in practice. Furthermore, we discover that the throughput of the K-means algorithm decreases with the number of iterations.
Therefore we simplify an efficient DPC algorithm with neither an iterative process nor more parameters, which outperforms other strategies on both accuracy and efficiency.

\section{Conclusion}
In this paper, we propose a token decoupling and merging method to simultaneously consider the token importance and diversity. 
Since token importance and diversity are orthogonal for token pruning, our method can be employed into exisiting token pruning methods to further improve the performance. We demonstrate that our method achieved the SOTA performance trade-off between accuracy and FLOPs without imposing extra parameters. We hope that this paper, which incorporates token importance and diversity, will provide insights for the future work of pruning visual transformers.

{\small
\bibliographystyle{ieee_fullname}
\bibliography{egbib}
}






\end{document}